%% file: Template.tex
\documentclass{article}
\usepackage[preprint]{spconf,amsmath,graphicx}

\copyrightnotice{\copyright\ IEEE 2018}
\toappear{Extended version of the accepted submission to {\it Proc.\ ICIP 2018, Oct. 07-10, 2018, Athens, Greece}}

\usepackage[utf8]{inputenc} 
\usepackage[T1]{fontenc}    
\usepackage{hyperref}       
\usepackage{url}            
\usepackage{booktabs}       
\usepackage{amsfonts}       
\usepackage{nicefrac}       
\usepackage{microtype}      
\usepackage[nolist,nohyperlinks]{acronym}
\usepackage{color}
\usepackage{amsmath}
\usepackage{graphicx}
\usepackage{subcaption}
\usepackage[table,xcdraw]{xcolor}
\usepackage{mathtools}

\graphicspath{{imgs/}}

\DeclarePairedDelimiter\floor{\lfloor}{\rfloor}

\title{E\lowercase{ff}N\lowercase{et}: An Efficient Structure for Convolutional Neural Networks}
\twoauthors
  {Ido Freeman, Lutz Roese-Koerner}
	{Aptiv \\
	Wuppertal, Germany \\
	\ninept{\{first\_name.last\_name\}@aptiv.com}
}
  {Anton Kummert}
	{University of Wuppertal\\
	Department of Electical Engineering\\
	\ninept{kummert@uni-wuppertal.de}
}
\begin{document}
%
\maketitle
\begin{abstract}
With the ever increasing application of Convolutional Neural Networks to customer products the need emerges for models to efficiently run on embedded, mobile hardware. Slimmer models have therefore become a hot research topic with various approaches which vary from binary networks to revised convolution layers. We offer our contribution to the latter and propose a novel convolution block which significantly reduces the computational burden while surpassing the current state-of-the-art. Our model, dubbed EffNet, is optimised for models which are slim to begin with and is created to tackle issues in existing models such as MobileNet and ShuffleNet.
\end{abstract}
\begin{keywords}
convolutional neural networks, computational efficiency, real-time inference
\end{keywords}
%
\input{intro.tex}
\input{related.tex}
\input{method.tex}
\input{implementation.tex}
\input{experiments.tex}
\input{mobile_v2.tex}
\input{conclusions.tex}




\bibliographystyle{IEEEbib}
\bibliography{strings,refs}

\end{document}

%% file: intro.tex
\section{Introduction}
With recent industrial recognition of the benefits of Artificial Neural Networks to product capabilities, the demand emerges for efficient algorithms to run in real-time on cost-effective hardware. This contradicts, in a way, the almost parallel university research. While the latter enjoys a relative freedom in terms of execution cycles and hardware, the former is subjected to market forces and product requirements.


Over the years multiple papers proposed different approaches for real-time inference on a small hardware. One example is the pruning of trained networks \cite{babaeizadeh2016noiseout}, \cite{dong2017learning}, \cite{molchanov2016pruning}. Another is the fix-point conversion of $32bit$ networks to as far as binary models \cite{rastegari2016xnor}. A more recent approach concentrates on the interconnectivity of the neurons and the very nature of the vanilla convolution layers.

A vanilla convolution layer consists, in its core, of a four-dimensional tensor which is swiped over an input signal in the following format \textit{[rows, columns, channels in, channels out]}, resulting in a quadruple-component multiplication, thus scaling the computational cost by a four-fold factor.

As $3 \times 3$ convolutions are now a standard, they become a natural candidate for optimisation.
Papers as \cite{howard2017mobilenets} (MobileNet) and \cite{zhang2017shufflenet} (ShuffleNet) set to solve this issue by separating the computations along the different dimensions. Yet in their methods they leave two issues unaddressed. First, both papers report taking large networks and making them smaller and more efficient. When applying their models to slimmer networks, the results diverge. Second, both proposed models create an aggressive bottleneck \cite{szegedy2016rethinking} for data flow through the network. This kind of a bottleneck might prove insignificant in models of high redundancy yet, as our experiments show, it has a destructive effect on smaller models. 

We therefore propose an alternative constellation which retains most of the proportional decrease in computations while having little to no effect on the accuracies. We achieve this improvement by optimising data flow and neglecting practices which prove harmful in this unique domain. Our novel convolutional blocks allow us either to deploy larger networks to low-capacity hardware or to increase efficiency of existing models.

%% file: related.tex
\section{Related Work}
Much of the work in the field focuses on hyper-parameter optimisation. Algorithms from this class are rather general both in terms of target algorithm and optimisation objective. \cite{snoek2012practical} proposed a Bayesian optimisation framework for black-box algorithms as CNNs \footnote{Convolutional Neural Networks} and SVMs \footnote{Support Vector Machines} by maximising the probability of increasing the model's accuracy. This could be combined with multi-objective optimisation as in \cite{horn2016multi} to optimise computational complexity as well. These methods mostly work well when initialised properly and many are limited in their search space \cite{bergstra2012random}. Using reinforcement learning, \cite{zoph2016neural} trained an LSTM \footnote{Long Short-Term Memory} \cite{hochreiter1997long} to optimise hyper-parameters for improved accuracy and speed. This along with recent evolutionary methods \cite{real2017large} exhibits less limitations on the search space but complicates the development by requiring additional steps.

An additional approach consists of decreasing the size of large models in a post-processing manner. Papers such as \cite{babaeizadeh2016noiseout}, \cite{dong2017learning} and \cite{molchanov2016pruning} proposed pruning algorithms with a minimal cost in accuracies. Pruning, however, leads to several issues. The development pipeline requires an additional phase with dedicated hyper-parameters which require optimisation. Furthermore, as the network's architecture is changed, the models require additional fine-tuning.

A further method for post-processing compression is the fix-point quantisation of models to primitives smaller than the common $32bit$ floats \cite{chen2017fxpnet}, \cite{lin2016fixed}, \cite{lai2017deep} and the binary networks of \cite{rastegari2016xnor}. Quantised models, although much faster, consistently show decreased accuracies compared to their baselines and are thus less appealing.

Last and most similar to this work, papers as \cite{chollet2016xception}, \cite{howard2017mobilenets} and \cite{zhang2017shufflenet} revisited the very nature of the common convolution operator. This involves the dimension-wise separation of the convolution operator, as discussed in \cite{jaderberg2014speeding}. Here, the original operation is approximated using significantly less FLOPs. \cite{szegedy2016rethinking} separated the $3 \times 3$ kernels into two consecutive kernels of shapes $3 \times 1$ and $1 \times 3$. The MobileNet model \cite{howard2017mobilenets} took a step further and separated the channel-wise from the spatial convolution which is also only applied depthwise, see \autoref{fig:mobileBlock}. By doing so, a significant reduction in FLOPs was achieved while the majority of computations was shifted to the point-wise layers.
Finally, the ShuffleNet model \cite{zhang2017shufflenet} addressed the stowage of FLOPs in the point-wise layers by dividing them into groups in a similar way to \cite{krizhevsky2012imagenet}. This lead to a drastic reduction in FLOPs with a rather small toll on the accuracies, see Figure 1 in \cite{zhang2017shufflenet} and \autoref{fig:shuffleBlock}.

The diversity of methods shows that there are multiple ways to compress a CNN successfully. Yet most methods assume a large development model which is adjusted for efficiency. They thus commonly seem to reach their limits when applied to networks which are slim to begin with. As many embedded systems have a limited specification, models are normally designed within these limitations rather than optimising a large network. In such environments, the limitations of \cite{howard2017mobilenets} and \cite{zhang2017shufflenet} become clearer thus laying the base for our EffNet model which shows the same capacity even when applied to shallow and narrow models.

Finally, notice that the methods above are not mutually exclusive. For example, our model could also be converted to fix-point, pruned and optimised for the best set of hyper-parameters.

%% file: method.tex
\begin{figure}[h]
	\centering
	\begin{subfigure}[t]{0.14\textwidth}
		\includegraphics[width=\textwidth]{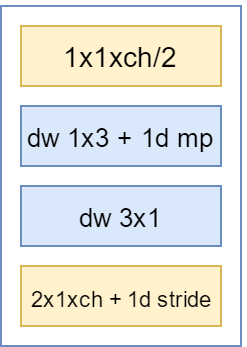}
		\caption{An EffNet block (ours)}
		\label{fig:ourBlock}
	\end{subfigure}
	~ 
	\begin{subfigure}[t]{0.145\textwidth}
		\includegraphics[width=\textwidth]{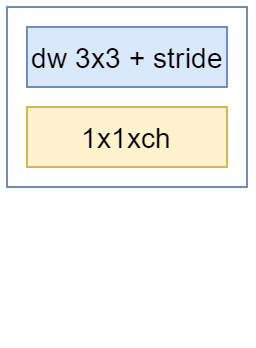}
		\caption{A MobileNet block}
		\label{fig:mobileBlock}
	\end{subfigure}
	~ 
	\begin{subfigure}[t]{0.14\textwidth}
		\includegraphics[width=\textwidth]{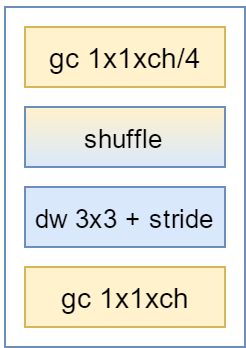}
		\caption{A ShuffleNet block}
		\label{fig:shuffleBlock}
	\end{subfigure}
	\caption{\ninept{A comparison of MobileNet and ShuffleNet with our EffNet blocks. 'dw' means depthwise convolution, 'mp' means max-pooling, 'ch' is for the number of output channels and 'gc' is for group convolutions. Best seen in colour.}}
	\label{fig:blockcompare}
\end{figure}
\vspace{-1.25em}

\section{Building Blocks for Increased Model Efficiency}
\label{sec:preliminary}
This section discusses the most common practices for increasing efficiency. The presented results assisted with identifying weaknesses in previous techniques and constructing a suitable solution in the form of a unified EffNet block. For practical reasons we avoid going into details regarding the exact settings of the following experiments. Instead we discuss their results and show their effect as a whole in \autoref{sec:experimets}.

The combination of multiple tasks, competitive costs and interactive run-times puts strict limitations on model sizes for industrial applications. These requirements in fact often lead to the use of more classical computer vision algorithms which are optimised to run a specific task extremely quick, e.g. \cite {tomasi1991detection}. Additionally, regulatory limitations often prohibit a one-network-solution as they require fallback systems and highly interpretable decision making processes.
Reducing the computational cost of all small classifiers in a project would thus allow either the redistribution of computational power to more critical places or enable deeper and wider models of larger capacity.

Exploring the limitations of previous work revealed that the smaller the model is, the more accuracy it loses when converted to MobileNet or to ShuffleNet, see \autoref{sec:experimets}. While analysing the nature of these suggested modifications we came across several issues.

\begin{table*}[t!!!!]
	\small
	\centering
	\caption{\ninept{Data flow in selected models. One could intuitively understand how an aggressive data compression in early stages would harm accuracies. Compression factors of $4$ or more are marked in red. $gc4$ means convolution in $4$ groups. Best seen in colour.}}
	\label{table:dataflow}
	\begin{tabular}{ll|ll|ll|ll}
		Baseline     &                                                       & MobileNet \cite{howard2017mobilenets}      &                                                       & ShuffleNet  \cite{zhang2017shufflenet}     &                                                       & EffNet (Ours)    &                                                       \\ \hline
		Layer        & \begin{tabular}[c]{@{}l@{}}Floats Out\end{tabular} & Layer           & \begin{tabular}[c]{@{}l@{}}Floats Out\end{tabular} & Layer           & \begin{tabular}[c]{@{}l@{}}Floats Out\end{tabular} & Layer            & \begin{tabular}[c]{@{}l@{}}Floats Out\end{tabular} \\ \hline
		3x3x64 + mp  & 16384                                                 & 3x3x64 + mp     & 16384                                                 & 3x3x64 + mp     & 16384                                                 & 1x1x32           & 32768                                                 \\
		&                                                       &                 &                                                       &                 &                                                       & dw 1x3 + 1d mp   & 16384                                                 \\
		& \textbf{}                                             &                 &                                                       &                 &                                                       & dw 3x1           & 16384                                                 \\
		&                                                       &                 &                                                       &                 &                                                       & 2x1x64 + 1d stride  & 16384                                                 \\ \hline
		3x3x128 + mp & 8192                                                  & dw 3x3 + stride & {\color[HTML]{FD6864} 4096}                           & gc4 1x1x32      & 8192                                                  & 1x1x64           & 16384                                                 \\
		&                                                       & 1x1x128         & 8192                                                  & dw 3x3 + stride & {\color[HTML]{FD6864} 2048}                           & dw 1x3 + 1d mp   & 8192                                                  \\
		&                                                       &                 &                                                       & gc4 1x1x128     & 8192                           & dw 3x1           & 8192                                                  \\
		&                                                       &                 &                                                       &                 &                                                       & 2x1x128 + 1d stride & 8192                                                  \\ \hline
		3x3x256 + mp & 4096                                                  & dw 3x3 + stride & {\color[HTML]{FD6864} 2048}                           & gc4 1x1x64      & 4096                                                  & 1x1x128          & 8192                                                  \\
		&                                                       & 1x1x256         & 4096                                                  & dw 3x3 + stride & {\color[HTML]{FD6864} 1024}                           & dw 1x3 + 1d mp   & 4096                                                  \\
		&                                                       &                 &                                                       & gc4 1x1x256     & 4096                                                  & dw 3x1           & 4096                                                  \\
		&                                                       &                 &                                                       &                 &                                                       & 2x1x256 + 1d stride & 4096                                                  \\ \hline
		Fully Connected           & {\color[HTML]{FD6864} 10}                             & Fully Connected              & {\color[HTML]{FD6864} 10}                             & Fully Connected              & {\color[HTML]{FD6864} 10}                             & Fully Connected               & {\color[HTML]{FD6864} 10}                            
	\end{tabular}
\end{table*}

\vspace{-1.25em}
\paragraph*{The Bottleneck Structure}
The bottleneck structure as discussed in \cite{iandola2016squeezenet} applies a reduction factor of eight to the number of input channels in a block w.r.t the number of output channels. A ShuffleNet block uses a reduction factor of four \cite{zhang2017shufflenet}. Yet narrow models do not tend to have enough channels for such a drastic reduction. In all of our experiments we witnessed a loss in accuracy comparing to a more moderate reduction. We therefore propose to use a bottleneck factor of two. Additionally it was found fruitful to use the spatial convolution (see following paragraph) with a depth multiplier of two, i.e. the first depthwise convolution layer also doubles the amount of channels.

\vspace{-1.25em}
\paragraph*{Strides and Pooling}
Both MobileNet and ShuffleNet models apply a stride of two to the depthwise spatial convolution layer in their blocks. Our experiments show two issues with this practice. First, we repeatedly witnessed a decrease in accuracy comparing to max-pooling. This was in a way expected as strided convolution is prone to aliasing.

Additionally, applying max-pooling to the spatial convolution layer does not allow the network to encode the data properly before it is reduced to a fourth of its incoming size. Nevertheless early stage pooling means cheaper following layers in a block. In order to maintain the advantages of early pooling while also relaxing data compression, we propose using separable pooling. Similar to separable convolution, we first apply a $2 \times 1$ pooling kernel (with corresponding strides) after the first spatial convolution layer. The second phase of the pooling then follows the last pointwise convolution of the block.

\vspace{-1.25em}
\paragraph*{Separable Convolutions}
Proposed by \cite{szegedy2016rethinking} but otherwise often neglected, we revisit the idea of consecutive separable spatial convolutions, i.e. using $3 \times 1$ and $1 \times 3$ layers instead of a single $3 \times 3$ layer. Separating the spatial convolution might only make a minor difference in terms of FLOPs but, combined with our pooling strategy it becomes more significant.

\vspace{-1.25em}
\paragraph*{Residual Connections}
Initially proposed by \cite{he2016deep} and quickly adopted by many, residual connections have become a standard practice.
Yet \cite{he2016deep} also showed that residual connections are mostly beneficial in deeper networks. We extend this claim and report a persistent decrease in accuracies throughout our experiments when using residual connections. We interpret this as a support to our claim that small networks cannot handle large compression factors well.

\vspace{-1.25em}
\paragraph*{Group Convolutions}
Following the promising results of \cite{zhang2017shufflenet}, we also experimented with similar configurations. The most drastic setting was the original ShuffleNet and the most relaxed one was the mere grouping of to the last point-wise layer in the blocks. The results showed a clear decrease in accuracies. We therefore refrained from using group convolutions, despite the appealing computational benefits.

\vspace{-1.25em}
\paragraph*{Addressing the First Layer}
Both MobileNet \cite{howard2017mobilenets} and ShuffleNet \cite{zhang2017shufflenet} avoided replacing the first layer. They claimed that this layer is rather cheap to begin with. We respectfully disagree and believe that every optimisation counts. After having optimised all other layers in the network, the first layer becomes proportionally larger. In our experiments, replacing the first layer with our EffNet block saves $\sim30\%$ of the computations for the respective layer.

%% file: implementation.tex
\section{The E\lowercase{ff}N\lowercase{et} Model}
\label{sec:method}

\subsection{Data Compression}
\label{subsec:data compression}
Analysing the effects of the various methods discussed in \autoref{sec:preliminary}, we established that small networks are very sensitive to data compression. Throughout the experiments, each practice which led to larger bottlenecks had also harmed the accuracies. For a better understanding of the data flow concept, \autoref{table:dataflow} lists the dimensionality of an input through the different stages of our Cifar10 \cite{krizhevsky2009learning} networks.

\subsection{The EffNet Blocks}
\label{subsec:the effnet blocks}
We propose an efficient convolutional block which both solves the issue of data compression and implements the insights from \autoref{sec:preliminary}. We design this block as a general construction to replace seamlessly the vanilla convolutional layers in, but not limited to, slim networks.

We start, in a similar manner to \cite{szegedy2016rethinking}, by splitting the $3 \times 3$ depthwise convolution to two linear layers. This  allows us to pool after the first spatial layer, thus saving computations in the second layer.

We then split the subsampling along the spatial dimensions. As seen in \autoref{table:dataflow} and in \autoref{fig:blockcompare} we apply a $1 \times 2$ max pooling kernel after the first depthwise convolution. For the second subsampling we choose to replace the common pointwise convolution with $2 \times 1$ kernels and a corresponding stride. This practically has the same amount of FLOPs yet leads to slightly better accuracies.

Following the preliminary experiments in \autoref{sec:preliminary}, we decide to relax the bottleneck factor for the first pointwise convolution. Instead of using one fourth of the output channels, we recognise a factor of $0.5$, with a minimal channel amount of $6$, as preferable.

%% file: experiments.tex
\section{Experiments}
\label{sec:experimets}
For the evaluation section, we selected datasets which comply with our general settings; a small number of classes and a relatively small input resolution. From the results of \cite{howard2017mobilenets} and \cite{zhang2017shufflenet}, which showed comparable accuracies to their baselines, we have no reason to believe that EffNet will perform significantly differently. We therefore focus on smaller models. For each dataset, we do a quick manual search for probable hyper-parameters for the baseline to fulfil the requirements; two to three hidden layers and small number of channels. The other architectures then simply replace the convolutional layers without changing the hyper-parameters.

Each experiment was repeated five times to cancel out the effects of random initialisation.

We used neither data augmentation nor pre-training on additional data as proposed by \cite{krause2016unreasonable}. Hyper-parameters were also not optimised as our goal was to replace the convolution layers in every given network with the EffNet blocks.

We used Tensorflow \cite{abadi2016tensorflow} and trained using the Adam optimiser \cite{kingma2014adam} with a learning rate of $0.001$ and $\beta_1 = 0.75$.

As complementary experiments, we evaluated a larger EffNet model with roughly the same amount of FLOPs as the baseline. This is dubbed \textit{large} in the following tables and comprises of a combination of two additional layers and more channels in \autoref{table:cifar10res} and \autoref{table:gtsrbres} or simply more channels in \autoref{table:svhnres}. We also trained versions of both ShuffleNet and MobileNet with more channels to match roughly the amount of FLOPs of our EffNet model thus evaluating comparability of the architectures.

\begin{table}[h]
	\small
	\centering
	\caption{\ninept{A model comparison on the Cifar10 dataset}}
	\label{comp_cifar10}
	\begin{tabular}{l|l|l|l}
						& Mean Accuracy     & Mil. FLOPs	& Factor          	\\ \hline
		Baseline  		& 82.78\%        	& 80.3          & 1.00             	\\ \hline
		EffNet large 	& \textbf{85.02\%}	& \textbf{79.8}	& \textbf{0.99}		\\ \hline
		\hline
		MobileNet  		& 77.48\%            & 5.8           & 0.07            	\\ \hline
		ShuffleNet 		& 77.30\%            & \textbf{4.7}  & \textbf{0.06} 	\\ \hline
		EffNet (ours)   & \textbf{80.20\%} 	 & 11.4          & 0.14  			\\ \hline
		MobileNet large & 78.18\%			 & 11.6			 & 0.14				\\ \hline
		ShuffleNet large& 77.90\%			 & 11.1			 & 0.14				\\ \hline
	\end{tabular}
	\label{table:cifar10res}
\end{table}

\vspace{-1.25em}

\subsection{Cifar10}
As a simple, fundamental dataset in computer vision, Cifar10 \cite{krizhevsky2009learning} is a good example of the sort of tasks we aim to improve on. Its images are small and represent a limited number of classes. We achieve a significant improvement over MobileNet and ShuffleNet while still requiring $\sim7$ times less FLOPs than the baseline (\autoref{comp_cifar10}). We relate this improvement to the additional depth of the network meaning that the EffNet blocks simulate a larger, deeper network which does not underfit as much as the other models.

\subsection{Street View House Numbers}
Similar to Cifar10, the SVHN benchmark \cite{netzer2011reading} is also a common dataset for evaluation of simple networks. The data consists of $32 \times 32$ pixel patches centred around a digit with a corresponding label. \autoref{comp_svhn} shows the results of this experiment which favour our EffNet model both in terms of accuracy and FLOPs.

\begin{table}[h]
	\small
	\centering
	\caption{\ninept{A model comparison on the SVHN dataset}}
	\label{comp_svhn}
	\begin{tabular}{l|l|l|l}
						& Mean Accuracy    & kFLOPs         	& Factor        \\ \hline
		Baseline   		& 91.08\%          & 3,563.5        	& 1.00          \\ \hline
		EffNet large 	& \textbf{91.12\%} & \textbf{3,530.7}	& \textbf{0.99}	\\ \hline
		\hline
		MobileNet  		& 85.64\%          & 773.4	        & 0.22          \\ \hline
		ShuffleNet 		& 82.73\%          & 733.1          & 0.21          \\ \hline
		EffNet (ours)  	& \textbf{88.51\%} & \textbf{517.6} & \textbf{0.14} \\ \hline
	\end{tabular}
	\label{table:svhnres}
\end{table}

\vspace{-1.25em}

\subsection{German Traffic Sign Recognition Benchmark}
A slightly older dataset which is nevertheless very relevant in most current driver assistance applications is the GTSRB dataset \cite{Stallkamp-IJCNN-2011}.
With over $50,000$ images and some 43 classes it presents a rather small task with a large variation in data and is thus an interesting benchmark. As even small networks started overfitting very quickly on this data, we resized the input images to $32 \times 32$ and used dropout \cite{srivastava2014dropout} with a drop-probability of $50\%$ before the output layer. Results are shown in \autoref{comp_gtsrb} and also favour our EffNet model.

\begin{table}[h]
	\small
	\centering
	\caption{\ninept{A model comparison on the GTSRB dataset}}
	\label{comp_gtsrb}
	\begin{tabular}{l|l|l|l}
						& Mean Accuracy    & kFLOPs         	& Factor        \\ \hline
		Baseline   		& 94.48\%          & 2,326.5        	& 1.00          \\ \hline
		EffNet large 	& \textbf{94.82\%} & \textbf{2,171.9}	& \textbf{0.93}			\\ \hline
		\hline
		MobileNet  		& 88.15\%          & 533.0          	& 0.23          \\ \hline
		ShuffleNet 		& 88.99\%          & 540.7          	& 0.23          \\ \hline
		EffNet (ours)  	& \textbf{91.79\%} & \textbf{344.1} 	& \textbf{0.15} \\ \hline
	\end{tabular}
	\label{table:gtsrbres}
\end{table}

\vspace{-1.25em}

%

%% file: mobile_v2.tex
\section{Comparison with MobileNet v2}
\label{sec:mobilev2}
As \cite{sandler2018inverted} came out at the same time as our work, we extend this work and write a quick comparison. Finally, we show how using a few minor adjustments, we surpass \cite{sandler2018inverted} in terms of accuracy while being similarly expensive to compute.

\subsection{Architecture Comparison}
Both \cite{sandler2018inverted} and this work separate the convolution operation along some of its dimensions to save computations. Unlike \cite{sandler2018inverted}, we also separate the spatial, two-dimensional kernels into two single-dimensional kernels. We did notice a small decrease in accuracies of around 0.5\% across our experiments by doing so, yet it allows for a significantly more efficient implementation and requires less computations.

For tackling the data compression problem, \cite{sandler2018inverted} proposes to inflate significantly the amount of data throughout their block by multiplying the number of channels of the input by a factor of 4 – 10. This makes the compression less aggressive comparing to the respective block’s input while also moving it to the end of the blocks, i.e. a reversed bottleneck. They further recognise an interesting, often overlooked property of the ReLU function. When following a Batch Normalisation layer, ReLU sets half of the data to zero thus further compressing the data. To counter the problem, \cite{sandler2018inverted} resolves to a linear pointwise convolution at the end of each block. In practice they get a linear layer followed by another non-linear pointwise layer, i.e. $B*(A*x)$ with $x$ being the input, $A$ the first layer and $B$ the second. Removing layer $A$ altogether simply forces the network to learn the layer $B$ as the function $B * A$. Our experiments also showed an indifference to the existence of layer $A$. Nevertheless, we show that using a leaky ReLU \cite{xu2015empirical}
 on top of the layer $A$ significantly increases the performance.

\subsection{EffNet Adaptations}
Considering latest experiments, we revise our architecture by introducing three minor adjustments.

First, considering the bottleneck structure, we define the output channels of the first pointwise layer as a function of the block’s input channels rather than its output channels. Similar to \cite{sandler2018inverted}, yet less extreme, the number of channels is given by
\begin{equation*}
\floor{\frac{inputChannels * expansionRate}{2}}
\end{equation*}

Second, the depth multiplier in the spatial convolution, which we only previously increased in some cases, is now natively integrated into our architecture and set to $2$.

Last, we replace the ReLU on the pointwise layers with a leaky ReLU.

Please note that the experiments were not decisive regarding both the activation function for the depthwise convolution and the first layer in the network. For the sake of simplicity, we used a ReLU with the remark that both leaky ReLU and linear spatial convolution were occasionally preferable. The first layer in the following experiments is a vanilla convolutional layer with max pooling.

\subsection{Experiments}
We use the same datasets as in \autoref{sec:experimets} but aim at a different kind of comparison. We now evaluate three models.
\begin{enumerate}
	\item Our revised EffNet model
	\item The original MobileNet v2 model
	\item Our proposed modifications to the MobileNet v2 model with pooling instead of strided convolution and leaky ReLU instead of the linear bottleneck. This is dubbed \textit{mob\_imp} (mobile improved) throughout the following tables.
\end{enumerate}

The models are evaluated with three different expansion rates: 2, 4 and 6 while \textit{mob\_imp} is only tested with expansion rate of 6.
Tables \ref{comp_cifar10_v2}, \ref{comp_svhn_v2} and \ref{comp_gtsrb_v2} show how our revised architecture performs favourably to \cite{sandler2018inverted} in most settings in terms of accuracy while having only marginally more FLOPs. Furthermore, although the \textit{mob\_imp} model outperforms our model, it is significantly more expensive to compute.

\begin{table}[h]
	\small
	\centering
	\caption{\ninept{A comparison of MobileNet v2 and EffNet on the Cifar10 dataset for various expansion rates}}
	\label{comp_cifar10_v2}
	\begin{tabular}{l|l|l|l|l}
		Ex. Rate 	&                  	& Mean Acc.    		& Mil. Flops    & Fact.  		\\ \hline
					& Baseline         	& 82.78\%          	& 80.3          & 1.00          \\ \hline
		6           & EffNet           	& \textbf{83.20\%} 	& 44.1 & 0.55 \\
					& MobileNet v2     	& 79.10\%          	& \textbf{42.0}          & \textbf{0.52}          \\ \hline
		4           & EffNet           	& \textbf{82.45\%} 	& 31.1 & 0.39 \\
					& MobileNet v2     	& 78.91\%          	& \textbf{29.2}          & \textbf{0.36}          \\ \hline
		2           & EffNet           	& \textbf{81.67\%} 	& 18.1 & 0.22 \\
					& MobileNet v2     	& 76.47\%          	& \textbf{16.4}          & \textbf{0.20}          \\ \hline
		6			& mob\_imp 			& 84.25\%          	& 44.0          & 0.55         	\\ \hline
	\end{tabular}
\end{table}

\vspace{-1.25em}

\begin{table}[h]
	\small
	\centering
	\caption{\ninept{A comparison of MobileNet v2 and EffNet on the SVHN dataset for various expansion rates}}
	\label{comp_svhn_v2}
	\begin{tabular}{l|l|l|l|l}
		Ex. Rate 	&                  	& Mean Acc.    		& kFlops    		& Fact.  		\\ \hline
					& Baseline         	& 91.08\%           & 3,563.5       	& 1.00          \\ \hline
		6           & EffNet           	& \textbf{87.80\%} 	& 2,254.8 	& 0.63 \\
					& MobileNet v2     	& 87.16\%           & \textbf{2,130.4}          	& \textbf{0.60}          \\ \hline
		4           & EffNet           	& \textbf{87.49\%} 	& 1,729.5 	& 0.49 \\
					& MobileNet v2     	& 86.93\%          	& \textbf{1,646.6}          	& \textbf{0.46}          \\ \hline
		2           & EffNet           	& \textbf{87.30\%} 	& 1,204.2 	& 0.34 \\
					& MobileNet v2     	& 86.71\%          	& \textbf{1,162.8}          	& \textbf{0.33}          \\ \hline
		6			& mob\_imp 			& 88.78\%          	& 2,506.7          	& 0.70         	\\ \hline
	\end{tabular}
\end{table}

\vspace{-1.25em}

\begin{table}[h]
	\small
	\centering
	\caption{\ninept{A comparison of MobileNet v2 and EffNet on the GTSRB dataset for various expansion rates}}
	\label{comp_gtsrb_v2}
	\begin{tabular}{l|l|l|l|l}
		Ex. Rate 	&                  	& Mean Acc.    		& kFlops    		& Fact.  		\\ \hline
					& Baseline         	& 94.48\%           & 2,326,5       	& 1.00          \\ \hline
		6           & EffNet           	& \textbf{93.74\%}		 	& 1,208.3 	& 0.51 \\
					& MobileNet v2     	& 92.82\%  & \textbf{1,159.2}          	& \textbf{0.50}          \\ \hline
		4           & EffNet           	& \textbf{92.30\%} 	& 956.4 	& 0.41 \\
					& MobileNet v2     	& 91.56\%          	& \textbf{934.9}          	& \textbf{0.40}          \\ \hline
		2           & EffNet           	& 90.40\%		 	& \textbf{704.5} 	& \textbf{0.30} \\
					& MobileNet v2     	& \textbf{90.74\%} 	& 710.7          	& 0.31          \\ \hline
		6			& mob\_imp 			& 93.25\%          	& 1,408.0          	& 0.61         	\\ \hline
	\end{tabular}
\end{table}

\vspace{-1.25em}

%% file: conclusions.tex
\section{Conclusions}
We have presented a novel convolutional block for CNNs, called EffNet, which promises to reduce computational effort significantly while preserving and even surpassing the baseline's accuracy. Our unified block is designed to ensure the safe replacement of the vanilla convolution layers in applications for embedded and mobile hardware. As networks are reduced to a small fraction of the baseline's FLOPs, our method presents a two-fold advantage, first is the quicker inference and second the application of a larger, deeper network becoming possible. We have also shown how such a larger network is clearly preferable to the baseline while requiring a similar amount of operations.